\documentclass[11pt]{article}

\usepackage[final]{acl}

\usepackage{times}
\usepackage{latexsym}

\usepackage[T1]{fontenc}

\usepackage[utf8]{inputenc}

\usepackage{microtype}

\usepackage{inconsolata}

\usepackage{graphicx}

\usepackage{algorithm}
\usepackage{algpseudocode}
\usepackage{amsmath}
\usepackage{amssymb} 

\usepackage{graphicx}
\usepackage{subcaption}   %
\usepackage{booktabs}     %
\usepackage{multirow}
\usepackage{makecell}

\usepackage{microtype}
\usepackage{xcolor}
\usepackage{soul}         %
\usepackage{verbatim}

\title{Eco-MobiLLM: Efficient LLM Fine-Tuning on the Mobile Device via Server Assisted Side-Tuning}

\author{
\textbf{
Xingke Yang$^{1,*}$,
Liang Li$^{2,*}$,
Zhiyi Wan$^{3}$,
Sicong Li$^{1}$,
Xiaoqi Qin$^{3}$,
Jiang Liu$^{4}$,
Tomoaki Ohtsuki$^{5}$
}
\\
\textbf{
Xin Fu$^{1}$,
Miao Pan$^{1}$
}
\\[0.5em]
$^{1}$University of Houston, USA \quad
$^{2}$Pengcheng Laboratory, China \quad
\\
$^{3}$Beijing University of Posts and Telecommunications, China \quad
\\
$^{4}$Waseda University, Japan \quad
$^{5}$Keio University, Japan
}

\begin{document}
\maketitle
\begingroup
\renewcommand{\thefootnote}{}
\footnotetext{$^{*}$Equal contribution}
\endgroup

\begin{abstract}
 There is a huge gap between numerous intriguing applications fostered by on-device large language model (LLM) fine-tuning (FT) from fresh mobile data and the limited resources of a mobile device. While existing server-assisted methods (e.g., split learning or side-tuning) may enable LLM FT on the local mobile device, they suffer from heavy communication burdens of activation transmissions. In addition, their multiple on-device forward passes over the same samples across training epochs lead to substantial iterative computation on the mobile device. To address those issues, we develop Eco-MobiLLM, an efficient LLM FT method which can be deployed on the mobile device via server-assisted side-tuning. To reduce communication cost, Eco-MobiLLM develops an activation shortcut that transmits only the tokens involved in the loss calculation instead of full activation matrices to guide the side network tuning. Besides, to improve computing efficiency and further accelerate FT convergence, Eco-MobiLLM introduces a novel two-stage training paradigm,  which allows the server to reuse historical activations and saves the mobile device from repeatedly computing forward passes for the recurring data samples. Extensive experimental results demonstrate Eco-MobiLLM's superiority.
\end{abstract}

\section{Introduction}


\textcolor{black}{Large language models (LLMs) like GPT~\cite{brown2020language} and LLaMA~\cite{touvron2023llama} have recently demonstrated remarkable capabilities across various NLP tasks~\cite{ren2024representation,ye2024cross,brown2020language,hu2020xtreme} via pretraining on public corpora followed by domain-specific fine-tuning (FT). Although effective in centralized cloud environments, this paradigm faces fundamental challenges in mobile computing scenarios, where user data are continuously generated across distributed mobile devices, raw data upload raises privacy concerns, and intelligent services are subject to stringent latency demands~\cite{Li2021infocom,ye2020federated,kang2017neurosurgeon,tambe2021edgebert}. How to enable LLM FT on the mobile device \textcolor{black}{without sacrificing LLM performance} becomes an essential and popular research topic.}

Despite those promising applications, on-device LLM FT faces significant hardware  resource challenges. Because LLMs scale to billions of parameters, both the computational and memory requirements for fine-tuning an LLM can be prohibitively high for existing mobile devices' hardware. Beyond storing the full model parameters, the predominant LLM training algorithm necessitates substantial memory for intermediate activations, gradients, and optimizer states. For example, \textcolor{black}{even with parameter-efficient fine-tuning (PEFT) methods such as LoRA~\cite{hulora}}, fine-tuning a 1.3B-parameter LLM typically requires over 14.5 GB of GPU memory, exceeding the capacity of most mobile devices, which often have 4–12 GB of RAM~\cite{xu2024fwdllm}. A single training epoch for such an LLM could require hours of processing on mobile GPUs~\cite{cai2023efficient}, which means full on-device fine-tuning cycles will take multiple days. This memory usage and latency fundamentally conflict with practical deployment scenarios requiring continuous model adaptation.

\begin{figure}[!t]
	\centering
	\includegraphics[width=0.99\linewidth]{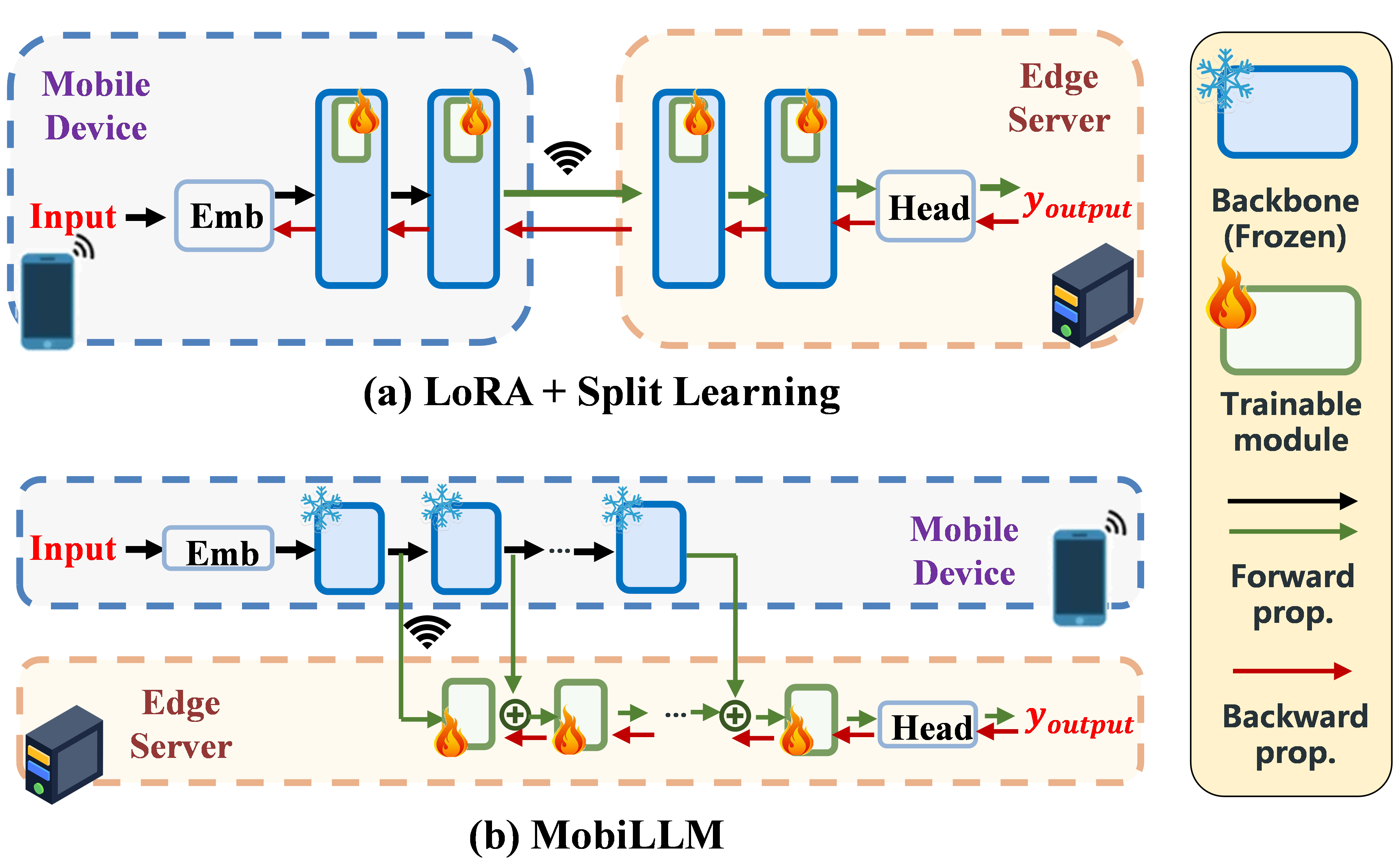}
	\caption{Device-server collaborative LLM FT.}
	\label{fig:SLandMobiLLM}
    \vspace{-4pt}  
\end{figure}

To enable LLM FT on a computing resource constrained mobile device, researchers are exploring to get some help from the high-performance computing server and collaborate the mobile device and the server via wireless communications supported by current mobile edge computing infrastructures. An intuitive solution is to integrate split learning~\cite{gupta2018distributed} with PEFT, i.e., training shallow layers on the mobile device while offloading deeper layers to the server as shown in Fig.~\ref{fig:SLandMobiLLM}(a). 
Yet, the sequential processing and bidirectional communications in this approach require fast wireless access and strict synchronization. Another device-server collaboration LLM FT approach is MobiLLM~\cite{li2025mobillm}, which introduces server-assisted parallel side-tuning. Instead of splitting the model vertically by layers, MobiLLM decouples the trainable modules (hosted on the server) from the frozen backbone model (retained on the mobile device) as shown in Fig.~\ref{fig:SLandMobiLLM}(b). 
By eliminating backpropagation, MobiLLM enables the LLM FT of billion-parameter models (e.g., OPT-1.3B) on the mobile device within 4.5 GB memory.

Although the pioneering device-server collaboration framework relieves the memory concerns of mobile devices, it still faces challenges for the practical deployment of LLM FT on the mobile device. (i) Extra communication burden of activation transmissions: Both split learning and MobiLLM have to transfer full-dimension activations/gradients iteratively to the server, which incurs extra communication burdens. Especially when processing large batches with long sequence inputs, the communication cost may be too high to execute LLM FT with low-speed access or time-varying channels in wireless edge networks. (ii) Iterative on-device computation cost during LLM FT: Existing device-server collaboration designs require the mobile devices to iteratively compute the activations across all training epochs. That induces enormous cumulative computation costs on the resource-constrained mobile device, which grows proportionally with LLM FT duration. \textcolor{black}{(iii) Limited convergence speed of LLM FT: In existing collaborative paradigm where the device remains involved throughout LLM FT, training time efficiency is limited by slow computation on the mobile device together with communication latency. For serial split learning, these two sources of delay are linearly accumulated. Although MobiLLM partially overlaps these costs through a parallel design, slow device-side computation still remains the dominant bottleneck in overall training speed.}


\textcolor{black}{To tackle those issues, in this paper, we propose Eco-MobiLLM, a communication-, computation-, and training-time-efficient LLM fine-tuning framework for mobile devices via device–server collaboration. Our contributions are summarized as follows: }


\noindent (i) We introduce a token selection method that transmits only critical backbone-derived activations to supervise the side-network training at the server, thereby significantly reducing the communication workload.

\noindent (ii)  \textcolor{black}{ We propose a novel two-stage training paradigm, where the mobile device performs computation and uploading only in the first epoch, and the subsequent training is completed on the server by reusing the data cached during the first epoch. This not only accelerates the side-network training on the server side, but also enables the device to skip redundant forward propagation on the frozen backbone for recurring samples, reducing both computational and transmission costs.} 



\noindent (iii) We implement Eco-MobiLLM on various mobile devices and evaluate its performance on sub-billion and billion-scale LLMs across diverse tasks and system settings. Experimental results demonstrate that compared to the second best LLM FT method on the mobile device, Eco-MobiLLM achieves at least a $1.79\times$ reduction in device-side computation, a $17.03\times$ reduction in communication cost and a $5.19\times$ reduction in convergence time, while maintaining competitive LLM fine-tuning performance.

\section{Preliminaries}

\subsection{Parameter- and Memory-Efficient FT}
The memory footprint of LLM fine-tuning arises from three core components: model parameters, intermediate activations, and optimizer states. Extensive research has focused on optimizing these aspects, with Parameter-efficient Fine-tuning (PEFT) methods emerging as a prominent approach to reduce optimizer state memory and gradient computation costs by selectively updating sparse subsets of parameters while maintaining performance comparable to full fine-tuning. Techniques such as Adapter~\cite{houlsby2019adapter}, which inserts bottleneck modules between transformer layers, LoRA~\cite{hulora}, which decomposes weight updates using low-rank matrices, and BitFit~\cite{zaken2021bitfit}, which exclusively trains bias terms, exemplify this paradigm. 
However, these approaches remain constrained by activation memory overhead as their trainable parameters remain embedded within the original model architecture, necessitating full backward passes through all layers. To address this limitation, recent innovations like ladder side-tuning (LST)~\cite{sung2022lst} and LoSA~\cite{mercea2024time} propose decoupling trainable components into a lightweight side network, creating a backpropagation highway that bypasses activation storage for the original model. 

\subsection{Device-Server Collaborative LLM FT}

\textcolor{black}{To enable device-side LLM adaptation that can exploit fresh, user-specific signals}, researchers are exploring on-device fine-tuning techniques. Given mobile hardware constraints, a common strategy involves partitioning the LLM and offloading part of the computation/storage burdens to adjacent servers or devices. Particularly, the split learning~\cite{gupta2018distributed} structure retains and updates only the few shallow transformer layers on the mobile device while delegating deeper layers training to servers, \textcolor{black}{thereby reducing on-device training cost and resource usage.} As shown in Fig.~~\ref{fig:SLandMobiLLM}(a), the training framework integrates split learning with LoRA in a device-server collaborative setup, where activations of split layers are transmitted during forward propagation, and corresponding gradients are exchanged during backpropagation.
When extended to multi-device scenarios, the LLM is further divided into layer-wise sub-models across devices, with forward/backward computations executed sequentially in a relay fashion~\cite{chen2023confidant}. However, this layer-wise partitioning enforces strict sequential execution, and requires each device to update partial trainable parameters. In contrast, MobiLLM~\cite{li2025mobillm} introduces an innovative framework that decouples LLM FT into a frozen backbone model and a trainable side network, where the server hosts the adapter-based side network parallel to the device-side backbone (Fig.~\ref{fig:SLandMobiLLM}(b)). This framework maintains memory-efficient forward passes and preserves the original LLM functionality on resource-constrained devices, while offloading computation-intensive backpropagation to high-performance servers. During LLM FT, this parallel device-server training allows the device to execute lightweight inference through the frozen backbone while the server simultaneously computes gradients via the side network.

\begin{table}[!t]
    \small
    \setlength{\tabcolsep}{2pt}
    \captionsetup{justification=centering}
    \caption{Communication and computational costs (On Xavier, task: MRPC@OPT-1.3B, batch size=8, seq len=256).}
    \label{table:comm_flops}
    \begin{tabular}{c|c|c|c}
\bottomrule
Method                 & \begin{tabular}[c]{@{}c@{}}Per-Batch\\ Comm. (MB)\end{tabular} & \begin{tabular}[c]{@{}c@{}}Per-Epoch\\ Comp. (PFLOPS)\end{tabular} & \begin{tabular}[c]{@{}c@{}}Total Comp.\\ (PFLOPS)\end{tabular} \\ \hline
BitFit-local   & -                                                             & 18.4                                                              & 208.3                                                          \\ \hline
BitFit-SL & 33.4                                                          & 1.5                                                               & 17.4                                                           \\ \hline
MobiLLM                & 200.4                                                         & 9.5                                                               & 124.5                                                          \\ \toprule
\end{tabular}
\end{table}


\begin{figure*}
    \centering
    \includegraphics[width=0.99\textwidth]{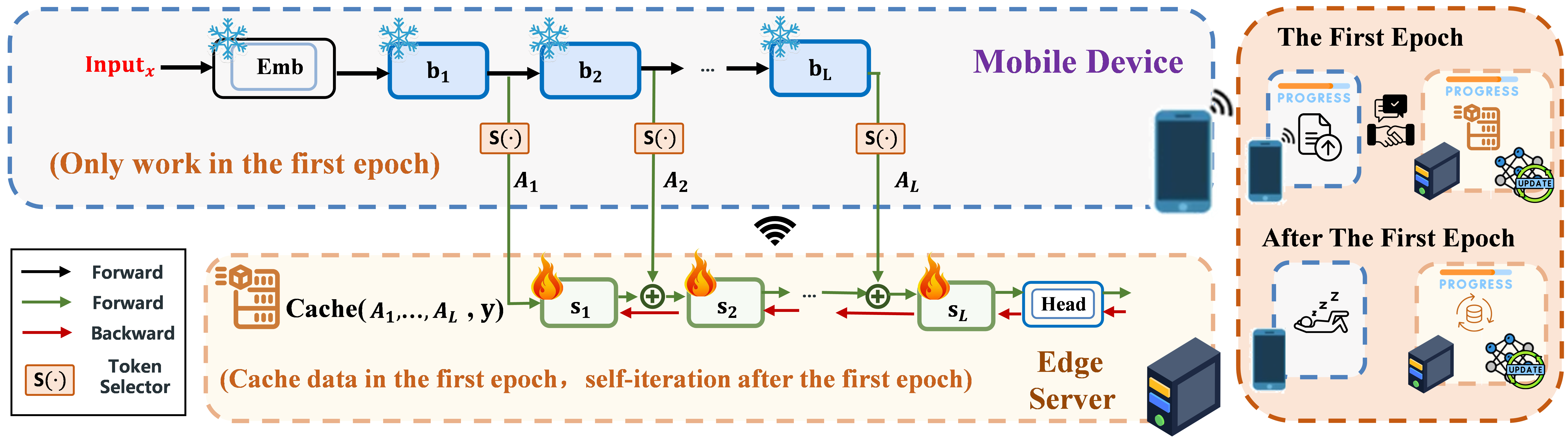}
    \caption{An overview of the Eco-MobiLLM system. }
    \label{fig:overview}
\end{figure*}

\section{Motivation}\label{sec:motivation}
While existing device-server collaboration methods enable LLM FT on the mobile device, there are still many critical issues for practical deployment in terms of computing efficiency, communication efficiency, and \textcolor{black}{training time efficiency}.

\textbf{Observation 1: High device-server communication burden.} Current collaborative LLM FT methods require frequent device-server communications for exchanging intermediate activations or gradients—bidirectional in split learning (activations and gradients) and unidirectional in MobiLLM (activations only). This results in substantial transmission loads, especially for large models processing long sequences or large batches. For instance, in split learning, fine-tuning OPT-1.3B with a batch size of 8 and token length of 256 requires transmitting up to 33.4 MB of data for just one batch (see Table~\ref{table:comm_flops}). Despite advanced future wireless accessing technologies (e.g., 6G), communications remain a critical issue given the exponential growth of LLM parameters and edge deployment demands. 

\textcolor{black}{\textbf{Observation 2: Iterative computation on the mobile device.} 
LLM FT tasks typically demand dozens to hundreds of training epochs for convergence, requiring iterative training on the same set of local samples. While device-server collaborative frameworks offload partial computations to the server, mobile devices remain burdened by iterative local computation. For instance, in the split learning framework, the mobile device must recompute and transmit the intermediate activations in every epoch—even for identical input samples—due to continuous parameter updates on the device-side model. MobiLLM eliminates device-side backpropagation by maintaining a frozen pretrained backbone, but it still enforces re-computation and transmission of static intermediate activations across training epochs by default.} Table~\ref{table:comm_flops} compares communication and computational costs of device-only BitFit, split-learning BitFit, and MobiLLM. Since the required number of epochs in a fine-tuning session usually depends on dynamic factors like dataset size and hyperparameter configurations, prolonging on-device participation could make resource planning on memory/energy-constrained devices inherently unstable and inefficient.

\section{Eco-MobiLLM Design}
\subsection{Eco-MobiLLM Overview}

\textcolor{black}{To address the observed issues, we propose Eco-MobiLLM, \textcolor{black}{a two-stage} and efficient LLM FT design for the mobile device. \textcolor{black}{Eco-MobiLLM follows the device-server collaboration framework but effectively improves both computational and communication efficiency and further accelerates the overall training process.} Briefly, as shown in Fig.~\ref{fig:overview}, Eco-MobiLLM deploys a frozen pre-trained model on the mobile device and lets it perform forward propagation only during the first epoch when batching over the local dataset, while allocating all trainable modules to the server. }

\textcolor{black}{In the first training stage (the first epoch), the mobile device performs forward propagation over local batches and forwards the activation values $(\mathbf{A}_1, \dots, \mathbf{A}_L)$ between the transformer layers and the training target $\text{Label}_y$ to the server. Before transmission, the activations are compressed by a token selector $S(\cdot)$ to reduce communication cost. In parallel with the device-side forward passes, the server updates the side-network parameters using the received (\(\mathbf{A}_1, \dots, \mathbf{A}_L , \textit{Label}_y \)) and caches these pairs to construct a side-tuning guidance database. The privacy concerns associated with transmitting intermediate activations are discussed in Appendix~\ref{app:Privacy}. In the second training stage, the mobile device is freed from the training process, and the server continues to train the side network iteratively until convergence by reusing the cached information from the first stage. We will detail the key components, their functionalities and the procedure of Eco-MobiLLM in the subsections below.}

\subsection{\textcolor{black}{Key Computing-Efficient Components}}

\subsubsection{Adapter Side-Tuning Network}

\begin{figure}[!t]
	\centering
	\includegraphics[width=0.98\linewidth]{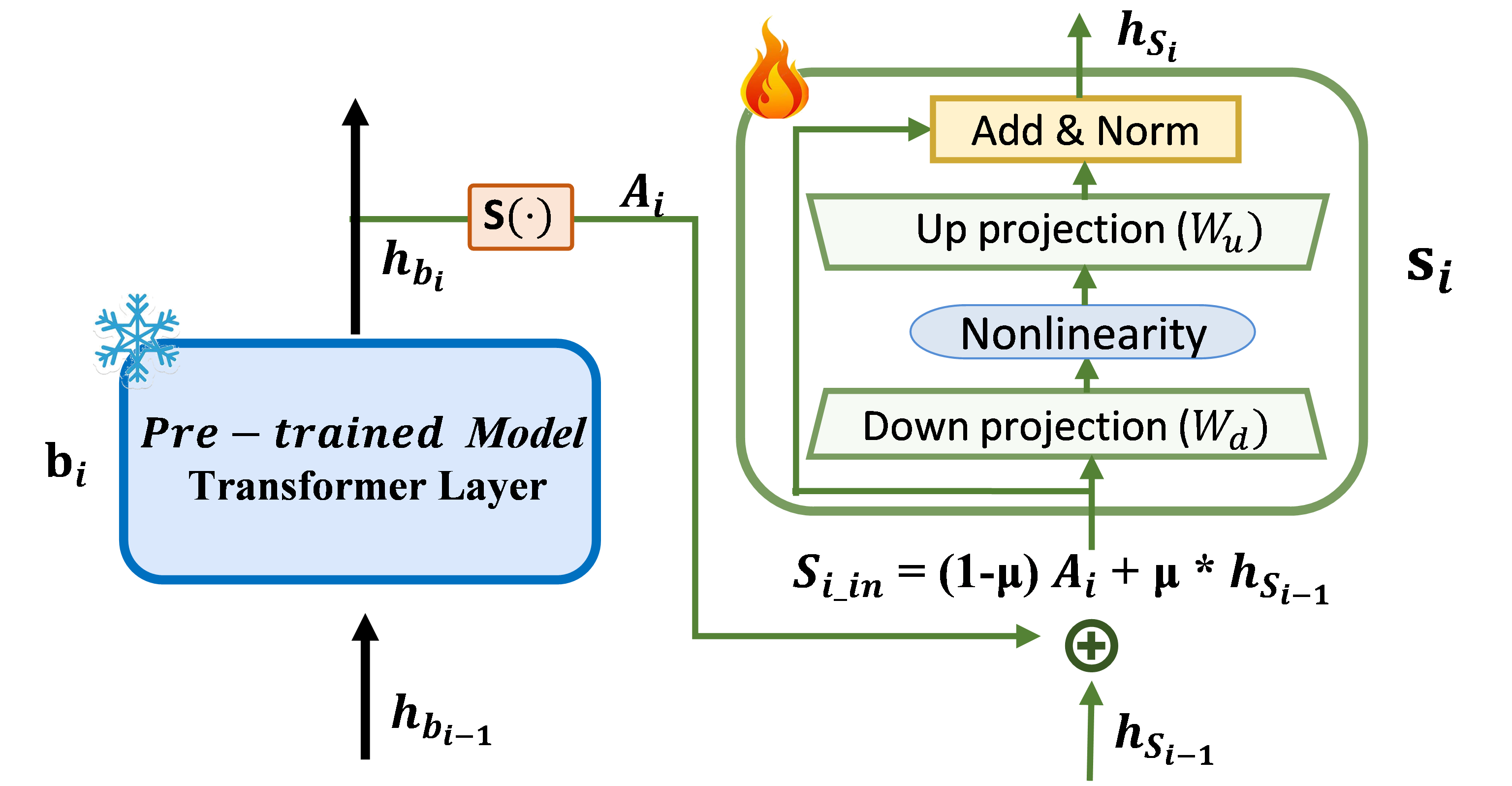}
	\caption{Sketches of the adapter side-tuning.}
	\label{fig:Fwcalculation}
    \vspace{-4pt} 
\end{figure}

\textcolor{black}{Aiming to improve computing efficiency, Eco-MobiLLM follows the device-server collaboration framework to offload the computing-intensive workload to the server. As shown in Fig. ~\ref{fig:overview},
Eco-MobiLLM lets the mobile device only retain the frozen pre-trained model for forward inference and offloads the training and updating of the side-network to the server. 
In Eco-MobiLLM, for a pre-trained model with \( L \) transformer layers, we construct \( L \) adapter modules, where the $i$-th adapter module corresponds to the $i$-th transformer layer. The input and output dimensions of each adapter module match the hidden size of the transformer layers, ensuring seamless optimization of the backbone's representations. The fine-tuning operations of the \( i \)-th layer are outlined in Fig. ~\ref{fig:Fwcalculation}.}

\textcolor{black}{During the forward propagation process, let the \( i \)-th layer of the backbone network be denoted by \( b_i \), and its hidden state output be represented by \( h_{b_i} \). Similarly, the \( i \)-th layer of the side network adapter is denoted by \( S_i \), and its output is represented by \( h_{S_i} \). The input \( S_{i\_in} \) to the \( S_i \) is formulated as}

\begin{equation}
     S_{i\_in} = (1 - \mu_i) \cdot A_i + \mu_i \cdot h_{S_{i-1}},
\end{equation}
\textcolor{black}{where \( A_i \) is the activation value selected from the hidden state of the backbone \( h_{b_i} \) by the token selector, \( h_{S_{i-1}} \) is the output of the previous layer of the side network, and \( \mu_i = \text{sigmoid}(\alpha_i) \) is a gating parameter, where \( \alpha_i \) denotes a learnable scalar initialized to zero.}

\textcolor{black}{Each adapter applies a down-projection matrix $\textbf{W}_{down}$$\in \mathbb{R}^{d*r}$ to reduce dimensionality, and then uses a non-linear activation function $\sigma(\cdot)$ and an up-projection matrix $\textbf{W}_{up}\in \mathbb{R}^{r*d}$ to restore the original dimension. Let \( d \) be the hidden dimension of the input and \( r \) denote the bottleneck dimension of the adapter. Here, \( d \) is inherited from the pre-trained model, and \( r \) is a tunable hyperparameter. The  \( i \)-th  adapter output includes a residual connection and is given by:}
\begin{equation}
    h_{S_i} = S_{i\_in} + \sigma(S_{i\_in} \cdot W_{\text{down}}) W_{\text{up}}.
\end{equation}
\textcolor{black}{After that, \( h_{S_i} \) is passed to the next adapter layer, where it will be further trained until the final fine-tuned representation of the side network is obtained. Since all trainable modules are sequentially connected in the side network pathway, backpropagation will be completely executed within the server.}

\textcolor{black}{Therefore, under Eco-MobiLLM collaborative network architecture, the mobile device only needs to take care of the forward pathway of the frozen backbone network based on the raw personal data, i.e., Input\textsubscript{x}, thereby avoiding the high computational cost of tunable parts in LLM FT.}

\subsubsection{\textcolor{black}{Two-Stage Training Paradigm}}\label{sec:Cache design}


Under the proposed Eco-MobiLLM framework, we observe that the backbone network on the mobile device and the side network on the server work independently, and the backbone network on the mobile device remains frozen, as shown in Fig.~\ref{fig:overview}. This means that for the same data samples, the intermediate activations generated by the backbone remain unchanged across iterations.

\textcolor{black}{Inspired by this observation, we develop a two-stage training paradigm that capitalizes on this architectural property. As shown in the rightmost part of Fig.~\ref{fig:overview}, during the first epoch, the mobile device and the server run in parallel. The mobile device continuously samples batches from its local training set, performs forward propagation and transmits the generated intermediate activations (\(\mathbf{A}_1, \dots, \mathbf{A}_L \)) and the training target \(  \textit{Label}_y \) to the server for one complete epoch. Meanwhile, the server trains the side network and caches the received activations and  \(  \textit{Label}_y \). After the first epoch, since Input\textsubscript{x} is the same and the backbone network is frozen on the mobile device, activations stay unchanged. The server will train the side network iteratively until its convergence based on the cached (\(\mathbf{A}_1, \dots, \mathbf{A}_L , \textit{Label}_y \)). Note that during activation collection, Eco-MobiLLM keeps the on-device frozen backbone in evaluation mode, disabling stochastic layers such as dropout so that cached activations remain deterministic and reusable across epochs. Stochastic regularization is still used for server-side side-network training. Under the two-stage paradigm, the mobile device only needs to compute the forward pass once and transmit the activation values to the server in the first epoch. This effectively improves computing efficiency and significantly reduces communication cost. In later epochs, the server reuses the cached data to iterate quickly, leading to a substantial speedup in LLM fine-tuning.}



\begin{figure}[!t]  
	\centering
	\includegraphics[width=0.99\linewidth]{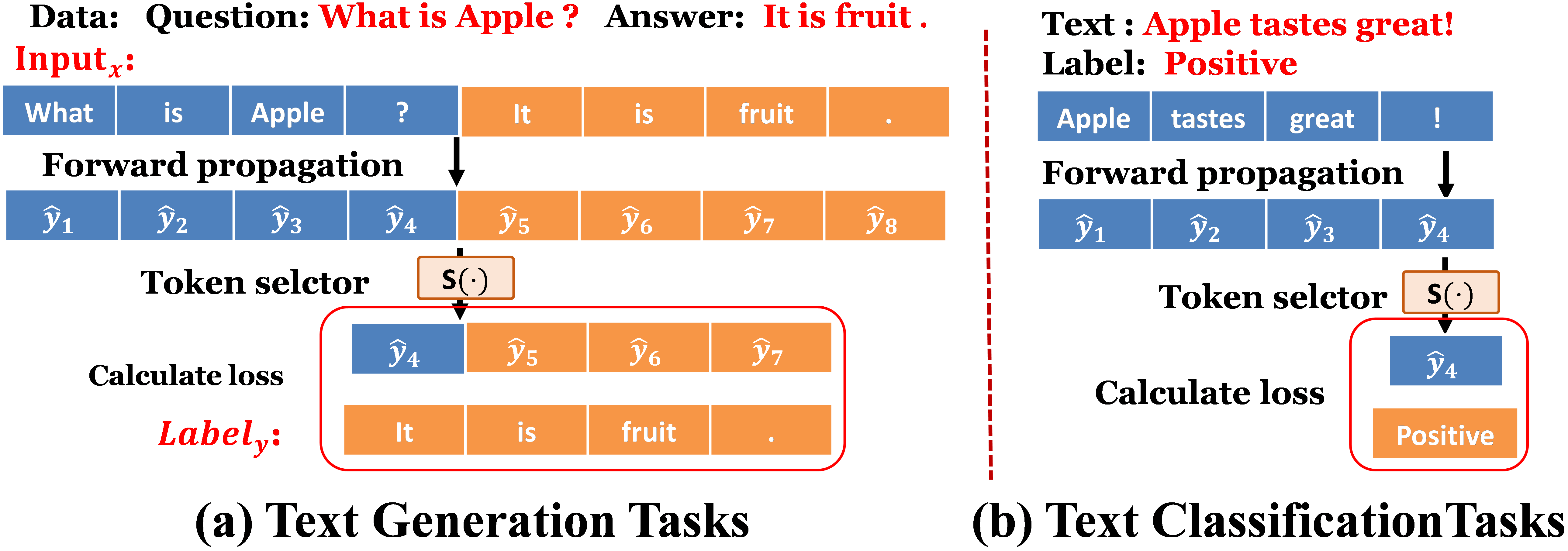}
	\caption{\textcolor{black}{Token-level selection for different LLM FT tasks.}}
	\label{fig:token flow}
    \vspace{-8pt} 
\end{figure}

\subsection{\textcolor{black}{Key Communication-Efficient Components}}\label{sec: 4.3}

\textcolor{black}{Although device-server collaboration approaches enable LLM FT on the mobile device, they introduce additional communication costs due to the transmission of activations. 
Aiming to minimize the communication cost, Eco-MobiLLM uses a token selector before activation transmissions.} 

\textcolor{black}{The token selector \( S(\cdot) \) is originally used in pre-trained models for the alignment of the loss calculations. As shown in Fig.~\ref{fig:token flow}, for FT training, the input sequence and the target label sequence often do not coincide in token counts. For the generation task, the input contains both the question and the answer concatenated as a complete semantic context, while the training target label sequence covers only the answer part. In classification tasks, each sentence corresponds to a single classification label, no matter how long it is. After forward propagation, to correctly calculate the supervised loss, it is necessary to align  \( \hat{y} \) with \(  \textit{Label}_y \) along the token dimension. Modern training procedures typically introduce a token-level selection function \( S(\cdot) \) to extract a subset from \( \hat{y} \) that matches the length of the label for calculating \( \textit{Loss}(\hat{y}, \textit{Label}_y) \).}

\textcolor{black}{We note that for the adapter modules in the side network, applying the computation first and then selecting the tokens yields the same result as selecting the tokens before the computation, because the adapter is applied independently to each token. Based on this, we propose applying the token selector \( S(\cdot) \), originally used in the output layer for loss computation, before transmission of intermediate activation values in each layer of the backbone network. By transmitting only the activations corresponding to token positions involved in loss computation, Eco-MobiLLM significantly reduces communication cost without sacrificing LLM FT performance.}

\section{Experimental Setup}

\textbf{Eco-MobiLLM Implementation:} \textcolor{black}{The experimental testbed consists of a server with NVIDIA A100 GPU and two representative mobile devices: (1) NVIDIA Jetson Xavier NX (8GB RAM, 4.6GB GPU memory available for training) and (2) Huawei MateBook laptop (16GB RAM). Xavier NX serves as the primary platform for evaluating the performance of GPU-accelerated LLM FT. The MateBook represents CPU-only scenarios for comparison under computation-constrained environments.}




\noindent\textbf{Models:} \textcolor{black}{To evaluate Eco-MobiLLM's performance, we employ two representative pre-trained architectures: the decoder-based OPT and encoder-based RoBERTa. Considering the Jetson Xavier platform's 4.6GB GPU memory constraint, we select four model variants: OPT-1.3B, OPT-350M, RoBERTa-large (350M), and RoBERTa-base (125M), ensuring architectural diversity while maintaining mobile compatibility. All models are initialized via Huggingface's Transformers library~\cite{wolf2019huggingface}.}


\noindent\textbf{Datasets:} \textcolor{black}{We take the GLUE benchmark~\cite{wang2018glue} and the DialogSum dataset ~\cite{chen2021dialogsum} for the evaluation of NLP tasks, which are widely used in the fine-tuning research for LLM~\cite{zhang2023fedpetuning,sun2024improving, sung2022lst}. GLUE benchmark comprises seven classification tasks and one regression task, including linguistic acceptability (CoLA ~\cite{warstadt2019neural}), sentiment analysis (SST-2~\cite{socher2013recursive}), similarity and paraphrase (MRPC~\cite{dolan2005automatically}, QQP~\cite{iyer2017first}, STS-B~\cite{cer2017semeval}), and natural language inference (MNLI~\cite{williams2017broad}, QNLI~\cite{rajpurkar2016squad}, RTE~\cite{bentivogli2009fifth}).} \textcolor{black}{DialogSum includes summaries of real-world conversations on a diverse set of topics and scenarios to evaluate text-generation tasks. \textcolor{black}{(See Appendix ~\ref{app:dataset} for dataset details.)}}

\noindent\textbf{Parameters:} \textcolor{black}{To ensure fair comparison, all experiments share the same configurations unless specified: FP16 precision (except FP32 for CPU-only devices), 20 training epochs, batch size 8, learning rate 5e-4 (except 1e-6 for full FT), the maximum sequence length 256, and averaged 60 Mbps in-lab Wi-Fi transmission speed. Additionally, the internal low-rank hidden size of LoRA, MobiLLM, and Eco-MobiLLM trainable modules is set to 64 by default.}

\begin{table*}[t]
\centering
\small
\setlength{\tabcolsep}{2.5pt}
\caption{Performance comparison of communication cost (Xavier, PBC: \underline{P}er \underline{B}atch \underline{C}omm. (MB), PEC: \underline{P}er \underline{E}poch \underline{C}omm. (GB), TC: \underline{T}otal \underline{C}omm. (GB)).}
\label{table:transmission}
\begin{tabular}{c|ccc|ccc|ccc|ccc}
\hline
Task
& \multicolumn{3}{c|}{MRPC@OPT-1.3B}
& \multicolumn{3}{c|}{DialogSum@OPT-1.3B}
& \multicolumn{3}{c|}{MRPC@RoBERTa-Large}
& \multicolumn{3}{c}{MRPC@RoBERTa-Base} \\ \hline

Target Acc. / ROUGE-1
& \multicolumn{3}{c|}{82.5 (Acc.)}
& \multicolumn{3}{c|}{20.9 (ROUGE-1)}
& \multicolumn{3}{c|}{87.1 (Acc.)}
& \multicolumn{3}{c}{85.4 (Acc.)} \\ \hline

\multirow{2}{*}{Method}
& \multicolumn{12}{c}{Communication Load} \\ \cline{2-13}
& \multicolumn{1}{c|}{PBC}
& \multicolumn{1}{c|}{PEC}
& \multicolumn{1}{c|}{TC}
& \multicolumn{1}{c|}{PBC}
& \multicolumn{1}{c|}{PEC}
& \multicolumn{1}{c|}{TC}
& \multicolumn{1}{c|}{PBC}
& \multicolumn{1}{c|}{PEC}
& \multicolumn{1}{c|}{TC}
& \multicolumn{1}{c|}{PBC}
& \multicolumn{1}{c|}{PEC}
& TC \\ \hline

Full FT-SL
& \multicolumn{1}{c|}{33.4}
& \multicolumn{1}{c|}{15.0}
& \multicolumn{1}{c|}{230.0}
& \multicolumn{1}{c|}{33.4}
& \multicolumn{1}{c|}{50.8}
& \multicolumn{1}{c|}{670.8}
& \multicolumn{1}{c|}{16.8}
& \multicolumn{1}{c|}{7.5}
& \multicolumn{1}{c|}{118.2}
& \multicolumn{1}{c|}{12.5}
& \multicolumn{1}{c|}{5.6}
& 82.2 \\ \hline

LoRA-SL
& \multicolumn{1}{c|}{33.4}
& \multicolumn{1}{c|}{15.0}
& \multicolumn{1}{c|}{171.8}
& \multicolumn{1}{c|}{33.4}
& \multicolumn{1}{c|}{50.8}
& \multicolumn{1}{c|}{645.3}
& \multicolumn{1}{c|}{16.8}
& \multicolumn{1}{c|}{7.5}
& \multicolumn{1}{c|}{83.4}
& \multicolumn{1}{c|}{12.5}
& \multicolumn{1}{c|}{5.6}
& 63.2 \\ \hline

BitFit-SL
& \multicolumn{1}{c|}{33.4}
& \multicolumn{1}{c|}{15.0}
& \multicolumn{1}{c|}{168.8}
& \multicolumn{1}{c|}{33.4}
& \multicolumn{1}{c|}{50.8}
& \multicolumn{1}{c|}{625.1}
& \multicolumn{1}{c|}{16.8}
& \multicolumn{1}{c|}{7.5}
& \multicolumn{1}{c|}{81.9}
& \multicolumn{1}{c|}{12.5}
& \multicolumn{1}{c|}{5.6}
& 57.0 \\ \hline

MobiLLM
& \multicolumn{1}{c|}{200.4}
& \multicolumn{1}{c|}{89.8}
& \multicolumn{1}{c|}{1174.1}
& \multicolumn{1}{c|}{200.4}
& \multicolumn{1}{c|}{304.9}
& \multicolumn{1}{c|}{3689.3}
& \multicolumn{1}{c|}{100.6}
& \multicolumn{1}{c|}{45.1}
& \multicolumn{1}{c|}{598.4}
& \multicolumn{1}{c|}{37.7}
& \multicolumn{1}{c|}{16.9}
& 217.5 \\ \hline

\textbf{Eco-MobiLLM}
& \multicolumn{1}{c|}{\textbf{0.78}}
& \multicolumn{1}{c|}{\textbf{0.35}}
& \multicolumn{1}{c|}{\textbf{0.35}}
& \multicolumn{1}{c|}{\textbf{24.1}}
& \multicolumn{1}{c|}{\textbf{36.7}}
& \multicolumn{1}{c|}{\textbf{36.7}}
& \multicolumn{1}{c|}{\textbf{0.39}}
& \multicolumn{1}{c|}{\textbf{0.17}}
& \multicolumn{1}{c|}{\textbf{0.17}}
& \multicolumn{1}{c|}{\textbf{0.15}}
& \multicolumn{1}{c|}{\textbf{0.07}}
& \textbf{0.07} \\ \hline

\end{tabular}
\end{table*}

\noindent\textbf{Baselines:}
We compare Eco-MobiLLM with three baseline paradigms: i) Device only fine-tuning (denoted by "-L"): Local execution using standard methods. ii) Split Learning (denoted by ``-SL’’): Integrates standard fine-tuning with split learning’s layer partitioning, executing the training pipeline serially across the device and the server. iii) MobiLLM: Server-assisted collaborative fine-tuning leveraging a side network to parallelize LLM adaptation. The compared methods are implemented as follows:

\noindent\textbf{Full-FT-L / Full-FT-SL}: Full model fine-tuning. \\
\textbf{LoRA-L/LoRA-SL~\cite{hulora}}: Inserts trainable low-rank matrices into the frozen backbone network.   \\
\textbf{BitFit-L/BitFit-SL~\cite{zaken2021bitfit}}: Updates only bias terms while freezing other weights.\\
\textbf{MobiLLM-L~\cite{li2025mobillm}}: Runs the frozen backbone and trains the side network locally on the mobile device. \\
\textbf{MobiLLM~\cite{li2025mobillm}}: Server-assisted side-tuning with full-activation transmission.

\section{Evaluation Results and Analysis}

\subsection{Advantages of Computing Efficiency}

\textcolor{black}{To evaluate the computational cost of the mobile device, we measure the total floating-point operations (FLOPs) required for LLM FT and use the MRPC task run to the target accuracy (Table~\ref{table:flops}) as a case study. \textcolor{black}{(See Appendix~\ref{app:ab_comp} for ablation results on the two components of Eco-MobiLLM.)} The results demonstrate: (1) server-assisted methods significantly reduce on-device computation compared to fully local LLM FT; (2) Eco-MobiLLM's \textcolor{black}{two-stage training paradigm} eliminates iterative on-device computations. Compared with MobiLLM, Eco-MobiLLM achieves a $13.1\times$ reduction in terms of on-device computing FLOPs, where the computing workload reduction is proportional to the number of training epochs. It indicates Eco-MobiLLM's \textcolor{black}{ one-time forward pass on-device computing design} yields the highest computing efficiency for the mobile device across all LLM FT schemes.}

\begin{table}[!t]
\small
\setlength{\tabcolsep}{3pt}
\caption{On-device computational cost (Xavier, task: MRPC).}
\label{table:flops}
\begin{tabular}{c|cc|cc}
\hline
Model                    & \multicolumn{2}{c|}{OPT-1.3B}                         & \multicolumn{2}{c}{RoBERTa-Large}       \\ \hline
Target Acc.             & \multicolumn{2}{c|}{82.5\%}                                & \multicolumn{2}{c}{87.1\%}                    \\ \hline
\multirow{2}{*}{Method} & \multicolumn{4}{c}{On-device Comp. (PFLOPs)} \\ \cline{2-5} 
                        & \multicolumn{1}{c|}{\begin{tabular}[c]{@{}c@{}}Per Epoch  \\ Comp.\end{tabular}}   
                        & \multicolumn{1}{c|}{\begin{tabular}[c]{@{}c@{}}Total   \\ Comp.\end{tabular}}   
                        & \multicolumn{1}{c|}{\begin{tabular}[c]{@{}c@{}}Per Epoch  \\ Comp.\end{tabular}}   
                        & \begin{tabular}[c]{@{}c@{}}Total   \\ Comp.\end{tabular}    \\ \hline
Full-FT-L               & \multicolumn{1}{c|}{28.5}   & \multicolumn{1}{c|}{439.1} & \multicolumn{1}{c|}{7.3}   & 114.4    \\ \hline
LoRA-L                  & \multicolumn{1}{c|}{20.5}   & \multicolumn{1}{c|}{235.5}  & \multicolumn{1}{c|}{5.2}   & 57.6    \\ \hline
BitFit-L                & \multicolumn{1}{c|}{18.4}   & \multicolumn{1}{c|}{208.3}  & \multicolumn{1}{c|}{4.9}   & 52.9    \\ \hline
MobiLLM-L               & \multicolumn{1}{c|}{9.7}   & \multicolumn{1}{c|}{127.3}  & \multicolumn{1}{c|}{2.5}   & 33.6    \\ \hline
Full-FT-SL              & \multicolumn{1}{c|}{2.4}   & \multicolumn{1}{c|}{36.7}  & \multicolumn{1}{c|}{0.6}   & 9.9    \\ \hline
LoRA-SL                 & \multicolumn{1}{c|}{1.6}   & \multicolumn{1}{c|}{18.4}   & \multicolumn{1}{c|}{0.4}   & 4.7    \\ \hline
BitFit-SL               & \multicolumn{1}{c|}{1.5}   & \multicolumn{1}{c|}{17.4}   & \multicolumn{1}{c|}{0.4}   & 4.3    \\ \hline
MobiLLM                 & \multicolumn{1}{c|}{9.5}   & \multicolumn{1}{c|}{124.5}    & \multicolumn{1}{c|}{2.4}   & 32.5    \\ \hline
\textbf{Eco-MobiLLM}    & \multicolumn{1}{c|}{9.5} & \multicolumn{1}{c|}{\textbf{9.5}} & \multicolumn{1}{c|}{2.4} & \textbf{2.4} \\ 
\hline
\end{tabular}
\vspace{-8pt} 
\end{table}

\subsection{Advantages of Communication Efficiency}

\textcolor{black}{As the results shown in Table~\ref{table:transmission}, compared to device-server collaborative methods, Eco-MobiLLM reduces at least $17.03\times$ communication cost (Eco-MobiLLM vs BitFit-SL, the OPT-1.3B model, and Dialogsum task). \textcolor{black}{(See Appendix~\ref{app:ab_comm} for ablation results of Eco-MobiLLM.)} We illustrate how Eco-MobiLLM improves communication efficiency from the following two aspects.}

\begin{figure*}[t]
\centering
\small
\subcaptionbox*{MRPC@OPT-1.3B (Xavier)\label{subfig:mrpc-opt1.3b-xavier}}{%
    \includegraphics[width=0.24\textwidth]{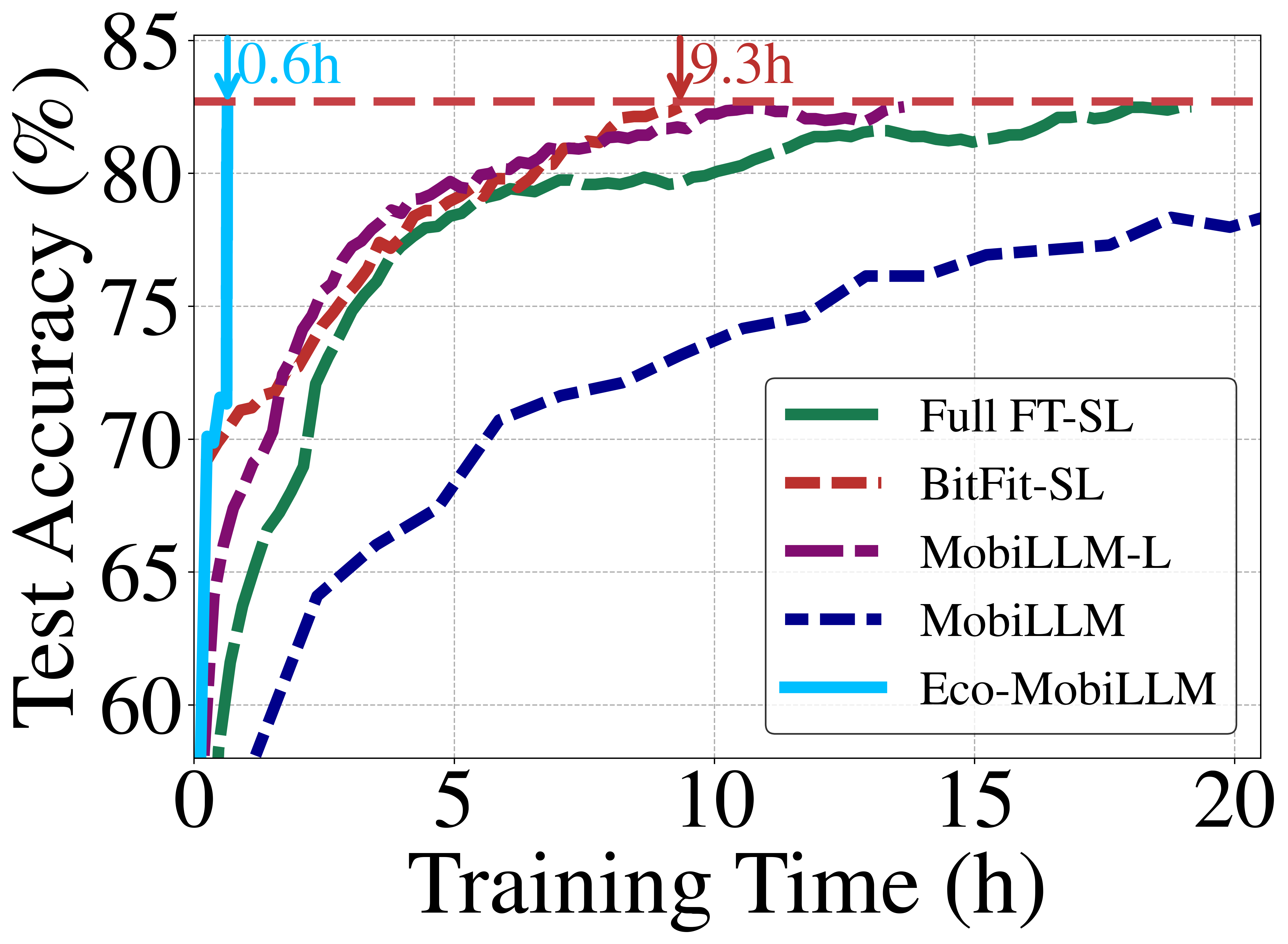}%
}\hfill
\subcaptionbox*{MRPC@OPT-1.3B (laptop)\label{subfig:mrpc-opt1.3b-laptop}}{%
    \includegraphics[width=0.24\textwidth]{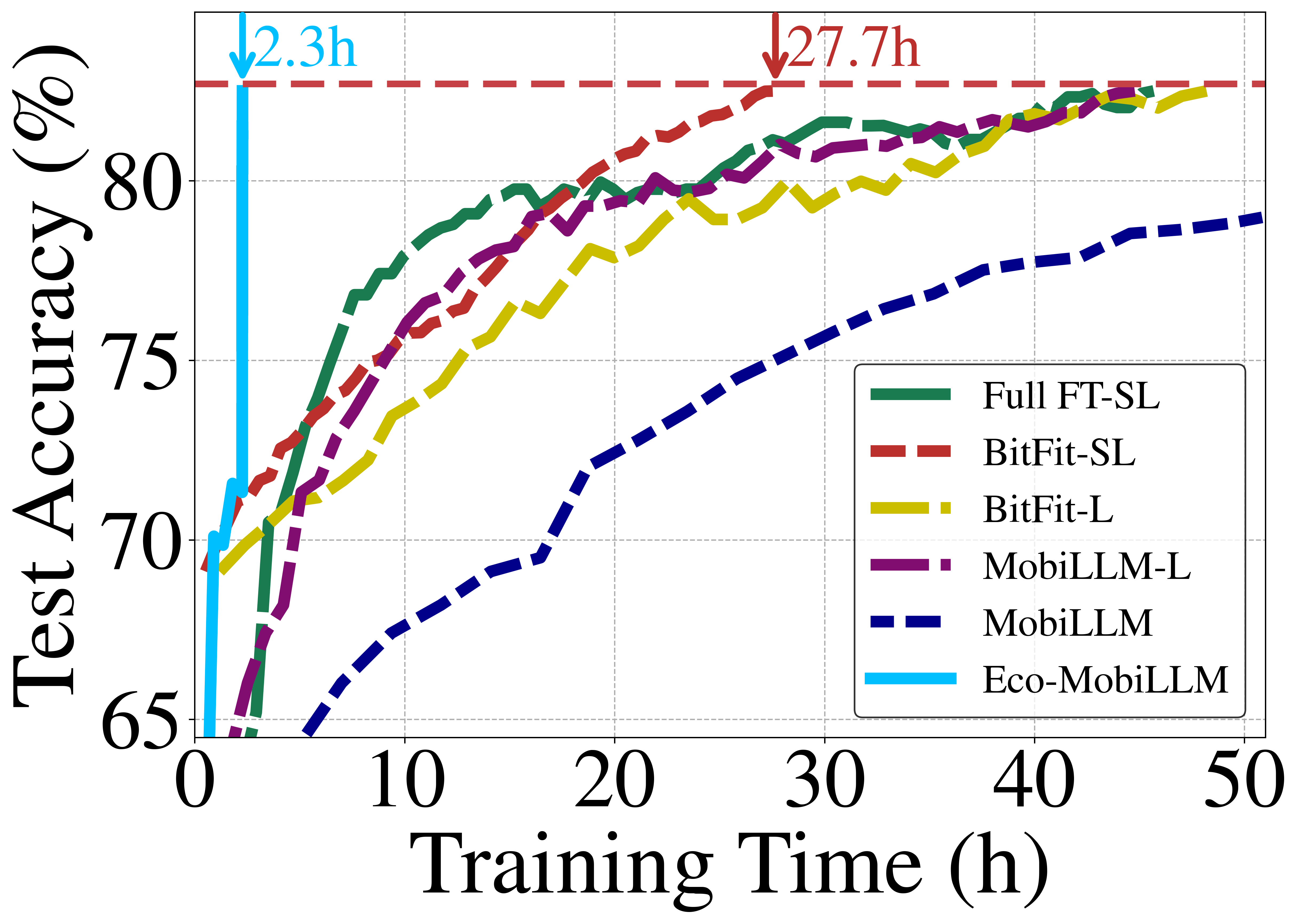}%
}\hfill
\subcaptionbox*{QNLI@RoBERTa-Base (Xavier)\label{subfig:qnli-robertab-xavier}}{%
    \includegraphics[width=0.24\textwidth]{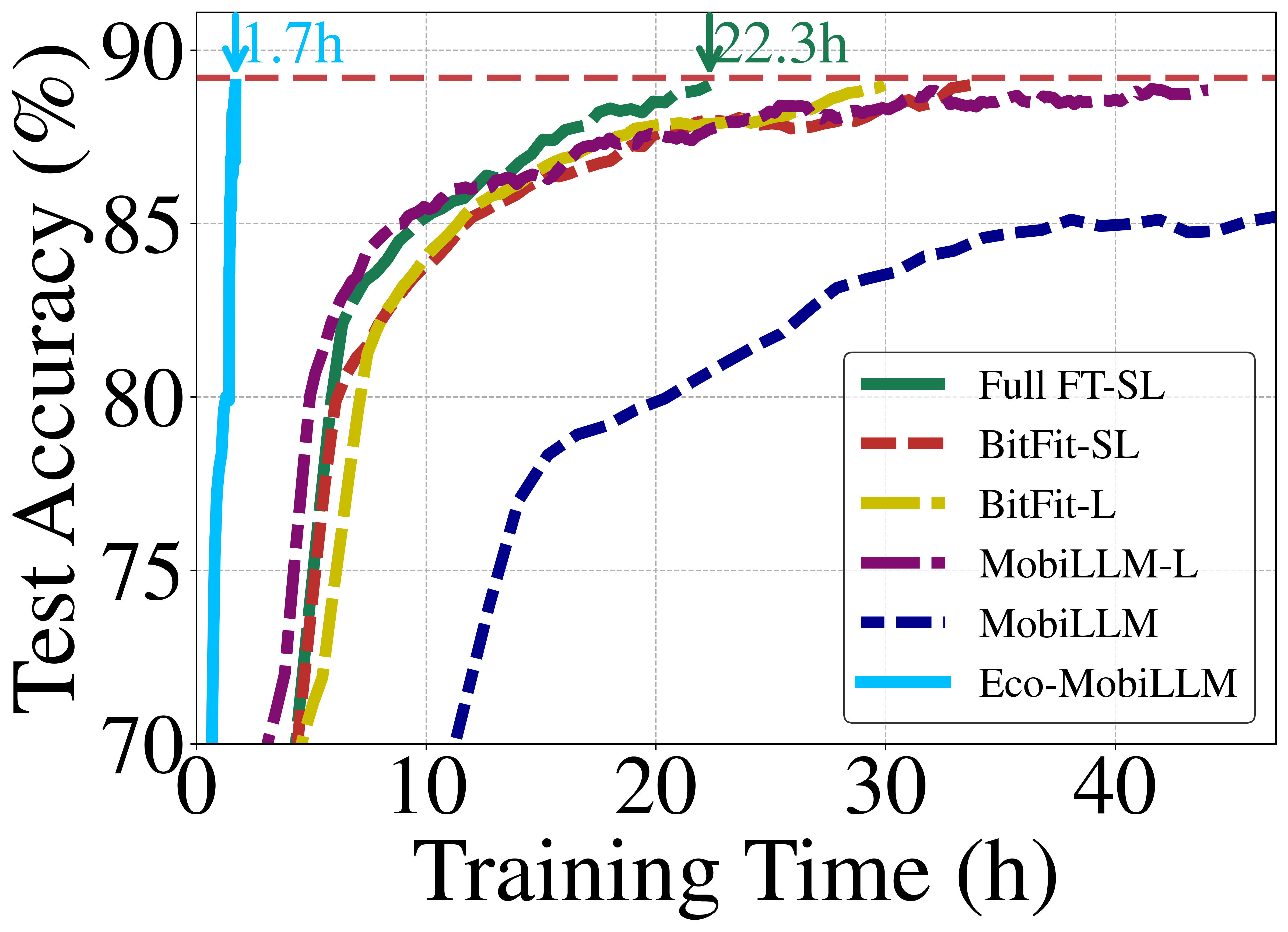}%
}\hfill
\subcaptionbox*{QNLI@RoBERTa-Base (laptop)\label{subfig:qnli-robertab-laptop}}{%
    \includegraphics[width=0.24\textwidth]{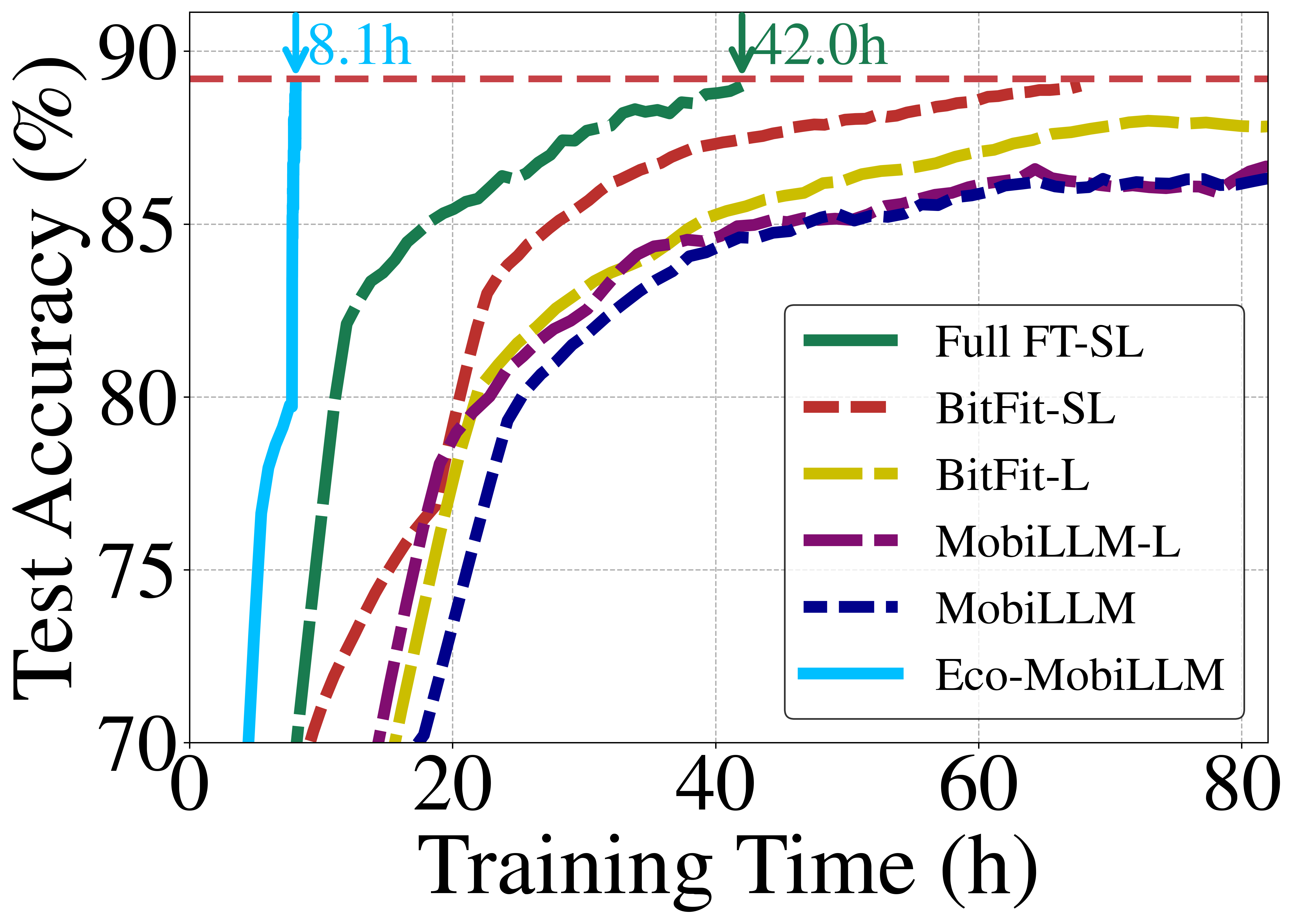}%
}%
\caption{Convergence performance comparisons over various models, tasks, and devices.}
\label{Fig:ConvOPTXavier}
\end{figure*}

\noindent\textbf{Per-Batch Transmission Efficiency.} Per-batch transmission measurements for OPT-1.3B (Table~\ref{table:transmission}) reveal fundamental architectural differences. Split learning based methods necessitate dual activation-gradient exchanges and result in 33.4MB/batch communication load, while MobiLLM further escalates this burden due to its full-layer activation transmissions. In contrast, Eco-MobiLLM leverages the token selector to identify and transmit only the activation dimensions involved in the loss computation. \textcolor{black}{Compared to split learning, it reduces per-batch traffic by at least 27.8\% in generation task and even up to 97\% in classification tasks, which makes such device-server collaboration based LLM FT feasible in low-speed or unstable wireless networks.}  

\noindent\textcolor{black}{\textbf{Total Communication Load.} As demonstrated in the Table~\ref{table:transmission}, all the other methods require iterative interactions with the server throughout the LLM FT process and lead to accumulated communication load proportional to the number of epochs. By contrast, Eco-MobiLLM processes the entire training dataset within the first epoch. It has no need for subsequent activation transmissions and only has a fixed initial cost. This promisingly enables precise estimation of total bandwidth requirements, which is critical for distributed training orchestration. }

\subsection{Advantages of LLM FT Acceleration}

\textcolor{black}{We evaluate the training efficiency of Eco-MobiLLM by measuring the total time required to reach target accuracy across diverse configurations involving two model architectures (OPT, RoBERTa), two tasks (MRPC, QNLI), and two mobile devices (NVIDIA Xavier, CPU-only laptop), as shown in Fig.~\ref{Fig:ConvOPTXavier}. For clarity, we simplify the presentation by retaining only BitFit's training curves (due to its similarity to LoRA) and excluding memory-intensive local FT baselines that exceed the mobile device's memory capacity.} \textcolor{black}{(See Appendix~\ref{app:ab_time} and ~\ref{app:time_breakdown} for the ablation study results and execution-time breakdown.)}

Eco-MobiLLM is much faster than conventional split learning and server-assisted LLM FT baselines. Taking the FT of the MRPC task with the OPT-1.3B model (on Xavier) as an example, Eco-MobiLLM is $15.5\times$ faster than the second-fastest approach, BitFit-SL. Particularly, Eco-MobiLLM demonstrates a two-phase acceleration pattern throughout the FT process. In the initial phase, Eco-MobiLLM’s parallel scheduling design allows the mobile device to process subsequent data batches while the server simultaneously trains side networks using current-batch activations. This eliminates the idle time of the mobile device/server during activation transmissions and server computations, which is a delay bottleneck in traditional split learning frameworks. Besides, using token selector before transmission in Eco-MobiLLM remarkably reduces the transmission delay, which further speeds up the training. 
\textcolor{black}{After the first epoch, Eco-MobiLLM leverages server-side caching to bypass redundant device-side computation and communication for identical sample activations. By leveraging the server's computational resources to perform independent iterative training, Eco-MobiLLM achieves faster LLM FT while maintaining stable convergence, which outperforms iterative recomputation and communication based LLM FT approaches.}

\subsection{LLM FT Performance Analysis}

\textcolor{black}{Eco-MobiLLM introduces two components beyond the MobiLLM framework, i.e., the token selector and \textcolor{black}{the two-stage training paradigm}.  We conducted ablation studies to assess the contribution of each component to LLM FT performance. For the GLUE benchmark, we report the average score, which combines accuracy on MNLI, QQP, QNLI, SST-2, MRPC, and RTE, Pearson correlation on STS-B, and Matthews correlation on CoLA. For the DialogSum dataset, we use ROUGE scores (R1/R2/RL) as the evaluation metric. As shown in Table~\ref{table:Eco_mobillm_efficiency}, \textcolor{black}{both the token selector and the two-stage training paradigm consistently maintain competitive performance across different datasets. (See Appendix~\ref{app:capacity_ablation} for additional fine-tuning performance results.)} The experimental results demonstrate that the proposed Eco-MobiLLM significantly improves training efficiency without sacrificing LLM FT performance.}


\begin{table}[!t]
\centering
\small
\setlength{\tabcolsep}{2pt}
\caption{Ablation study on LLM fine-tuning performance (TS: \underline{T}oken \underline{S}elector, TSTP: \underline{T}wo \underline{S}tage \underline{T}raining \underline{P}aradigm, w/o. : Without).}
\label{table:Eco_mobillm_efficiency}
\begin{tabular}{c|c|c|c}
\hline
Model & 
\multicolumn{2}{c|}{OPT-1.3B} & 
RoBERTa-Large \\ \hline
 \begin{tabular}[c]{@{}c@{}}Data\\ (Metrics)\end{tabular}
& \begin{tabular}[c]{@{}c@{}}GLUE\\ (Avg.)\end{tabular} 
& \begin{tabular}[c]{@{}c@{}}Dialogsum\\ (R1/R2/RL)\end{tabular} 
& \begin{tabular}[c]{@{}c@{}}GLUE\\ (Avg.)\end{tabular} \\ \hline

Eco-MobiLLM & 81.1 & 21.1 / 7.9 / 17.1 &  86.8 \\ \hline
w/o. TS & 81.2 & 21.1 / 7.7 / 17.0 &  86.7  \\ \hline
w/o. TSTP  & 81.1 & 21.0 / 7.7 / 17.1 &  86.8  \\ \hline
w/o. TS/TSTP & 81.4 & 21.3 / 7.6 / 17.5 & 86.6\\ 
\hline
\end{tabular}
\end{table}


\section{Conclusion}

\textcolor{black}{This paper has presented Eco-MobiLLM, a novel and efficient approach for mobile-device LLM fine-tuning via server-assisted side-tuning.} By decoupling the frozen backbone model on the mobile device from server-side side-network tuning, Eco-MobiLLM offloads computation and memory-intensive workloads to the server while enabling practical LLM FT on mobile devices. To improve communication efficiency, Eco-MobiLLM exploits token selection to compress activations and reduce transmission workloads. \textcolor{black}{To improve computing efficiency and further accelerate fine-tuning, Eco-MobiLLM introduces the two-stage training paradigm to remove iterative on-device computations by reusing historical activations.} Experimental results show that Eco-MobiLLM achieves at least a $1.79\times$ reduction in device-side computation, a $17.03\times$ reduction in communication cost, and a $5.19\times$ acceleration in LLM FT convergence, while maintaining competitive fine-tuning performance. Additional results on server-side cache storage, network-speed sensitivity, and transmitted-activation privacy are provided in Appendix~\ref{app:Discussion}.

\section*{Limitations}

\noindent \textcolor{black}{\textbf{Hardware-limited model-scale coverage:} Eco-MobiLLM focuses on more efficient deployment LLM FT on resource-constrained mobile devices, and our experiments are conducted on real edge platforms. Due to the limited device memory (e.g., on an NVIDIA Xavier with only 4.6 GB GPU memory available for training) and computing power, the largest model we can reliably run in our setting is 1.3B. However, we still evaluated multiple architectures (e.g., OPT and RoBERTa) and cover multi-scale evaluations (125M–1.3B) to provide reliable and representative results. Evaluating on a broader range of model sizes would be valuable, but our current evaluations also sufficiently demonstrate our work's efficiency advantages over existing methods in communication, computation, and training speed. As model sizes increase, other existing methods will face more severe issues with activation size and deployability.}

\noindent \textcolor{black}{\textbf{Fine-Tuning Data Setting:} This paper focuses on a standard on-device fine-tuning setting, where a mobile device fine-tunes the model over multiple epochs using a fixed local training dataset. This is also a common evaluation setting in existing mobile fine-tuning, PEFT, split-learning, and server-assisted fine-tuning studies. Under this setting, the same local samples are reused across epochs, so repeated device-side forward computation and activation transmission are exactly the bottlenecks we aim to address. Other research areas, such as continual learning~\cite{aljundi2019online,buzzega2020dark}, focus on settings in which new training data are continuously generated and evolve over time, which are beyond the primary scope of this work. Even in such setting, Eco-MobiLLM only needs to process each new sample once and upload its activations for later reuse, while other existing methods still need to repeatedly compute and transmit these activations in subsequent epochs.}

\bibliography{custom}

\newpage
\appendix

\section{Datasets Statistics} \label{app:dataset}



\begin{table}[!htbp]
\centering
\small
\caption{Dataset statistics.}
\label{tab:benchmarks}
\setlength{\tabcolsep}{2pt}
\begin{tabular}{l l l}
\hline
\textbf{Dataset} & \textbf{Description} & \textbf{\# Samples (train/eval)} \\
\hline
CoLA      & Linguistic Acceptability            & 8551 / 1043 \\
SST2      & Sentiment Analysis                  & 67349 / 872 \\
MRPC      & Sentence Equivalence                & 3668 / 408 \\
STSB      & Sentence Similarity                 & 5749 / 1500 \\
QQP       & Paraphrase Recognition              & 363846 / 40430 \\
RTE       & Textual Entailment                  & 2490 / 277 \\
QNLI      & Natural Language Inference          & 104743 / 5463 \\
MNLI      & Textual Entailment                  & 392702 / 9815 (9832) \\
DialogSum & Abstract Summary                    & 12460 / 500 \\
\hline
\end{tabular}
\end{table}

Table~\ref{tab:benchmarks} provides an overview of all the datasets utilized in our experiments, including task characteristics and sample counts.

\section{Further Experimental Results and Discussion}\label{app:Discussion} 

\subsection{Server Storage}\label{app:Server Storage}


\begin{table}[!htbp]
\centering
\small
\setlength{\tabcolsep}{4pt}
\caption{Server cache size across tasks (OPT-1.3B).}
\label{tab:server_cache_size}
\setlength{\tabcolsep}{10pt}
\begin{tabular}{l|c|c|c}
\hline
Method & \multicolumn{3}{c}{Server cache size (GB)} \\
\cline{2-4}
 & RTE & SST2 & QNLI \\
\hline
Eco-MobiLLM & 0.23 & 6.21 & 36.77 \\ \hline
w/o. TS      & 58.88 & 1589.76 & 3637.07 \\
\hline
\end{tabular}
\end{table}


The major focus of device-server collaborative LLM fine-tuning works is to reduce computation and communication costs on resource-constrained mobile devices. Most state-of-the-art studies~\citep{li2025mobillm, thapa2022splitfed, xu2024fwdllm} assume that the server has sufficient resources. We follow a similar setting and prioritize the analysis of task performance and system efficiency from the mobile-device side in the main paper.

Here, we provide quantitative measurements of Eco-MobiLLM’s server-side storage usage across different dataset scales (from RTE to QNLI). As shown in Table~\ref{tab:server_cache_size}, under the 1.3B model setting, the server-side cache is about 36.77 GB for a larger dataset such as QNLI (about 100K examples), while it is only about 0.23 GB for a small-scale dataset such as RTE (about 2K examples). Besides, the results also show that, without the token selector, enabling server-side caching is infeasible. Even for the smallest RTE dataset, it would require roughly 58 GB of cache. Consequently, although server caching offers clear benefits for on-device computation and training speed, it is impractical without the substantial reduction of uplink traffic from the token selector. This underscores the complementary nature of the two designs in Eco-MobiLLM.

Since device uploads only a small amount of selected activations, the storage overhead introduced by server-side caching does not make Eco-MobiLLM impractical, supporting its deployability. Given current mobile memory constraints and the fact that per-user datasets are usually relatively limited, these results indicate that, in our target scenario, the server-side storage cost is acceptable and does not raise serious feasibility concerns for the current design. However, when targeting larger backbones or higher data throughput, edge servers with tight storage budgets may require additional operational mechanisms, such as cache lifecycle management, eviction policies, and compression. We view these as meaningful directions for future work.



\subsection{Impact of Wireless Transmission Speed}
\label{app:wireless_speed}

\begin{figure}[!htbp]
\centering
\begin{subfigure}{0.23\textwidth}
    \centering    \includegraphics[width=\linewidth,height=0.8\linewidth]{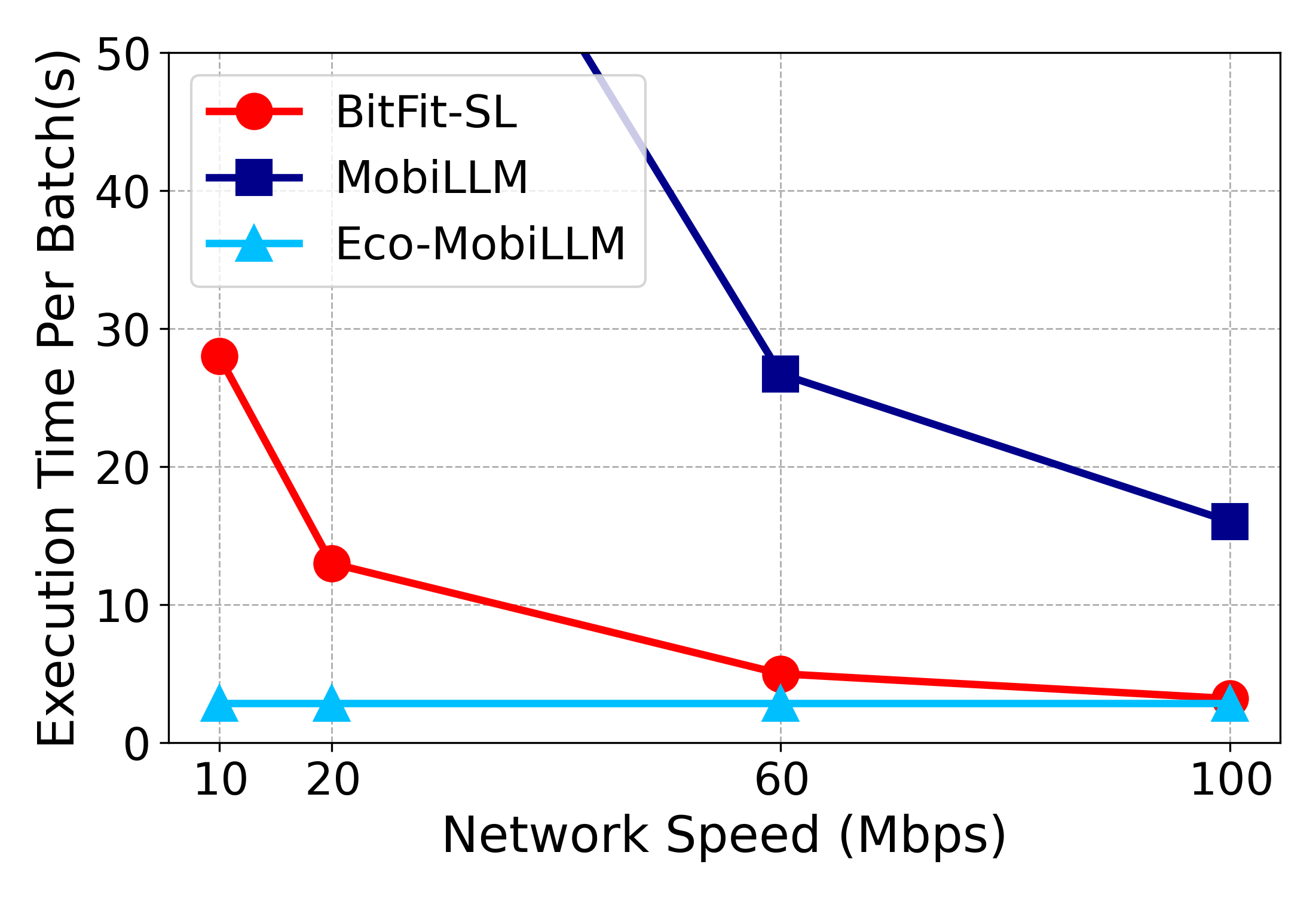}
    \caption*{\small (a) Task: SST-2} 
    \label{fig:sub1}
\end{subfigure}
\hfill 
\begin{subfigure}{0.23\textwidth}
    \centering    \includegraphics[width=\linewidth,height=0.8\linewidth]{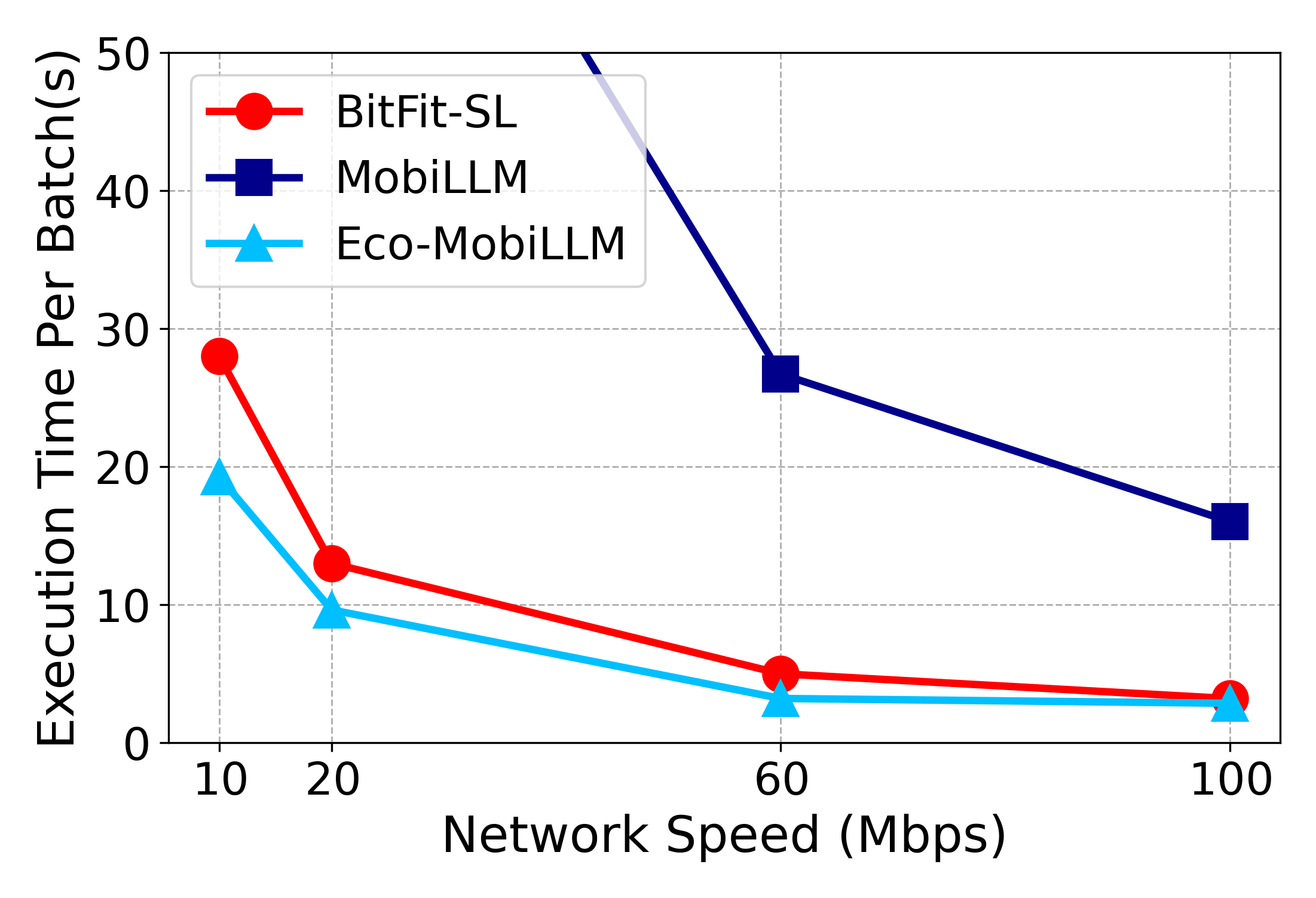}
    \caption*{\small (b) Task: DialogSum}
    \label{fig:sub2}
\end{subfigure}
\caption{\textcolor{black}{Impact of network speed on execution time per batch (On Xavier, OPT-1.3B).}}
\label{fig:time-netspeed}
\vspace{-6pt}
\end{figure}


Figure~\ref{fig:time-netspeed} illustrates the per-batch execution time across methods under different wireless transmission speeds. Since the total execution time in BitFit-SL and MobiLLM is predominantly affected by the communication time, they suffer severe performance degradation under low-speed transmission conditions. For example, BitFit-SL shows an $8.75\times$ slowdown at 10Mbps vs. 100Mbps. In contrast, Eco-MobiLLM maintains the strongest latency robustness for both classification and generation tasks across different transmission speeds. Such resilience stems from two facts: 1) Partial token transmission takes very short time and thus minimizes the impact of wireless transmission speeds, and 2) parallelized scheduling overlaps communication phases with device-side computation, hiding transmission delays in on-device computation time that remains unaffected by transmission speed fluctuations.

\subsection{Ablation Study of Eco-MobiLLM}

This section presents ablation results for the two key components of Eco-MobiLLM with respect to on-device computation, communication cost, and LLM fine-tuning acceleration, corresponding to and supporting the analyses in the main paper.

\subsubsection{Ablation Study of On-Device Computation}\label{app:ab_comp}


\begin{table}[!htbp]
\small
\caption{Ablation studies on the proposed Eco-MobiLLM designs regarding on-device computational cost. (Xavier, task: MRPC; TS: \underline{T}oken \underline{S}elector, TSTP: \underline{T}wo \underline{S}tage \underline{T}raining \underline{P}aradigm, w/o. : Without).}
\label{table:ab_flops}
\setlength{\tabcolsep}{4pt}
\begin{tabular}{c|cc|cc}
\hline
Model                    & \multicolumn{2}{c|}{OPT-1.3B}                         & \multicolumn{2}{c}{RoBERTa-Large}       \\ \hline
Target Acc.             & \multicolumn{2}{c|}{82.5\%}                                & \multicolumn{2}{c}{87.1\%}                    \\ \hline
\multirow{2}{*}{Method} & \multicolumn{4}{c}{On-device Comp. (PFLOPs)} \\ \cline{2-5} 
                        & \multicolumn{1}{c|}{\begin{tabular}[c]{@{}c@{}}Per Epoch  \\ Comp.\end{tabular}}   
                        & \multicolumn{1}{c|}{\begin{tabular}[c]{@{}c@{}}Total   \\ Comp.\end{tabular}}   
                        & \multicolumn{1}{c|}{\begin{tabular}[c]{@{}c@{}}Per Epoch  \\ Comp.\end{tabular}}   
                        & \begin{tabular}[c]{@{}c@{}}Total   \\ Comp.\end{tabular}    \\ \hline
Eco-MobiLLM               & \multicolumn{1}{c|}{9.5} & \multicolumn{1}{c|}{9.5} & \multicolumn{1}{c|}{2.4} & 2.4    \\ \hline
w/o. TS                & \multicolumn{1}{c|}{9.5} & \multicolumn{1}{c|}{9.5} & \multicolumn{1}{c|}{2.4} & 2.4    \\ \hline
w/o. TSTP                 & \multicolumn{1}{c|}{9.5}   & \multicolumn{1}{c|}{124.5}    & \multicolumn{1}{c|}{2.4}   & 32.5    \\ \hline
w/o. TS/TSTP  & \multicolumn{1}{c|}{9.5}   & \multicolumn{1}{c|}{124.5}    & \multicolumn{1}{c|}{2.4}   & 32.5 \\ 
\hline
\end{tabular}
\end{table}

As shown in Table~\ref{table:ab_flops}, the efficiency of on-device computation in Eco-MobiLLM stems from the two-stage training paradigm. The device performs a single forward pass in the first epoch and avoids subsequent redundant computations, so the on-device workload does not scale with the number of epochs and remains predictable.

\subsubsection{Ablation Study of Communication Cost}\label{app:ab_comm}

\begin{table}[!htbp]
\small
\setlength{\tabcolsep}{2pt}
\caption{Ablation studies on the proposed Eco-MobiLLM designs regarding communication cost.}
\label{table:ab_transmission}
\begin{tabular}{c|ccc}
\hline
Task & \multicolumn{3}{c}{MRPC@OPT-1.3B} \\
\hline
Target Acc. & \multicolumn{3}{c}{82.5 (Acc.)} \\
\hline
\multirow{2}{*}{Method} & \multicolumn{3}{c}{Communication Load} \\ \cline{2-4}
& \multicolumn{1}{c|}{{\begin{tabular}[c]{@{}c@{}}Per Batch \\ Comm. (MB) \end{tabular}}}
& \multicolumn{1}{c|}{{\begin{tabular}[c]{@{}c@{}}Per Epoch \\ Comm. (GB) \end{tabular}}}
& {\begin{tabular}[c]{@{}c@{}}Total \\ Comm. (GB) \end{tabular}} \\ \hline
Eco-MobiLLM
& \multicolumn{1}{c|}{0.78}
& \multicolumn{1}{c|}{0.35}
& 0.35 \\ \hline
w/o. TS
& \multicolumn{1}{c|}{200.4}
& \multicolumn{1}{c|}{89.8}
& 89.8 \\ \hline
w/o. TSTP
& \multicolumn{1}{c|}{0.78}
& \multicolumn{1}{c|}{0.35}
& 4.6 \\ \hline
w/o. TS/TSTP
& \multicolumn{1}{c|}{200.4}
& \multicolumn{1}{c|}{89.8}
& 1174.1 \\
\hline
\end{tabular}
\end{table}

\begin{table*}[t]
    \centering
    \caption{Ablation study on fine-tuning performance with different internal low-rank hidden sizes of the trainable side-network modules. (task: DialogSum@OPT-1.3B).  }
    \label{tab:capacity_ablation}

    \small
    \setlength{\tabcolsep}{9pt}
    \renewcommand{\arraystretch}{1.35}

    \begin{tabular*}{\textwidth}{@{\extracolsep{\fill}}lcccc@{}}
        \hline
        \textbf{Method}
        & \shortstack[c]{\textbf{Hidden Size = 16}\\(R1 / R2 / RL)}
        & \shortstack[c]{\textbf{Hidden Size = 64}\\(R1 / R2 / RL)}
        & \shortstack[c]{\textbf{Hidden Size = 128}\\(R1 / R2 / RL)}
        & \shortstack[c]{\textbf{Hidden Size = 256}\\(R1 / R2 / RL)} \\
        \hline

        Eco-MobiLLM
        & \textbf{19.8} / 7.0 / 15.5
        & 21.1 / \textbf{7.9} / 17.1
        & 21.9 / 8.4 / 17.9
        & 21.6 / 8.3 / \textbf{18.1} \\

        w/o. TS
        & \textbf{19.8} / 7.1 / 15.5
        & 21.1 / 7.7 / 17.0
        & \textbf{22.1} / 8.3 / \textbf{18.1}
        & 21.5 / 8.1 / 18.0 \\

        w/o. TSTP
        & 19.6 / \textbf{7.4} / 15.6
        & 21.0 / 7.7 / 17.1
        & 21.9 / 8.3 / 17.9
        & \textbf{21.7} / \textbf{8.4} / 18.0 \\

        w/o. TS/TSTP
        & 19.6 / 7.2 / \textbf{15.8}
        & \textbf{21.3} / 7.6 / \textbf{17.5}
        & 21.8 / \textbf{8.5} / 18.0
        & \textbf{21.7} / \textbf{8.4} / \textbf{18.1} \\

        \hline
    \end{tabular*}
\end{table*}

As shown in Table~\ref{table:ab_transmission}, the token selector and the two-stage training paradigm jointly contribute to the communication efficiency of Eco-MobiLLM. Specifically, the token selector significantly reduces the communication cost per batch, thereby driving down the cumulative communication cost per epoch and throughout the entire training process. Meanwhile, the two-stage training keeps the total communication cost at the level of the first epoch, further reducing network communication cost and ensuring its predictability.

\subsubsection{Ablation Study of LLM FT Acceleration}\label{app:ab_time}

\begin{table}[t]
\small
\setlength{\tabcolsep}{7pt}
\caption{Ablation studies on the proposed Eco-MobiLLM designs regarding Execution Time.}
\label{table:ab_time}
\begin{tabular}{c|ccc}
\hline
Task & \multicolumn{3}{c}{MRPC@OPT-1.3B} \\
\hline
Target Acc. & \multicolumn{3}{c}{82.5 (Acc.)} \\
\hline
\multirow{2}{*}{Method} & \multicolumn{3}{c}{Execution Time} \\ \cline{2-4}
& \multicolumn{1}{c|}{{\begin{tabular}[c]{@{}c@{}}Per Batch  \\ Time (s) \end{tabular}}} & \multicolumn{1}{c|}{{\begin{tabular}[c]{@{}c@{}}Per Epoch  \\ Time. (h) \end{tabular}}} & {\begin{tabular}[c]{@{}c@{}}Total  \\ Time (h) \end{tabular}} \\ \hline
Eco-MobiLLM    & \multicolumn{1}{c|}{2.86} & \multicolumn{1}{c|}{0.36} & 0.62 \\ \hline
w/o. TS    & \multicolumn{1}{c|}{26.72}  & \multicolumn{1}{c|}{3.41}  & - \\ \hline
w/o. TSTP       & \multicolumn{1}{c|}{2.86} & \multicolumn{1}{c|}{0.36} & 4.77 \\ \hline
w/o. TS/TSTP  & \multicolumn{1}{c|}{26.72} & \multicolumn{1}{c|}{3.41} & 44.67 \\
\hline
\end{tabular}
\end{table}

As shown in Table~\ref{table:ab_time}, Eco-MobiLLM’s accelerated training is primarily driven by two factors. First, the token selector significantly reduces per-batch communication cost, thereby compressing the transmission latency of intermediate activations. By overlapping communication with the device-side forward pass, it shortens the training time per batch. In addition, two-stage training paradigm allows subsequent epochs to reuse cached results, further accelerating overall training.

\begin{table*}[t]
\small
\centering
\caption{Per-Batch Execution Time Breakdown (task:SST-2@OPT-1.3B).}
\label{table:execution_time}
\setlength{\tabcolsep}{7pt}
\begin{tabular}{c|c|c|c|c|c|c|c}
\hline
\textbf{Method} 
& \multicolumn{2}{c|}{\textbf{Device: Xavier (Weak GPU)}} 
& \multicolumn{2}{c|}{\textbf{Device: Laptop (CPU-only)}} 
& \textbf{\begin{tabular}[c]{@{}c@{}}t3: Server \\ Comp. (s)\end{tabular}} 
& \multicolumn{2}{c}{\textbf{T: Total time (s)}} \\
\cline{2-5} \cline{7-8}
& t1: Comm.(s) & t2: Comp. (s) 
& t1: Comm.(s) & t2: Comp. (s) 
& & Xavier & Laptop \\
\hline
LoRA-L       & -    & -    & -    & 27.51 & -    & -    & 27.51 \\ \hline
BitFit-L     & -    & -    & -    & 26.80 & -    & -    & 26.80 \\ \hline
MobiLLM-L    & -    & 4.32 & -    & 19.29 & -    & 4.32 & 19.29 \\ \hline
Full-FT-SL   & 4.45 & 0.73 & 8.90 & 4.30  & 0.15 & 5.33 & 13.35 \\ \hline
LoRA-SL      & 4.45 & 0.55 & 8.90 & 2.74  & 0.10 & 5.10 & 11.74 \\ \hline
BitFit-SL    & 4.45 & 0.53 & 8.90 & 2.69  & 0.09 & 5.07 & 11.68 \\ \hline
MobiLLM      & 26.72 & 2.85 & 53.44 & 10.30 & 0.01 & 26.72 & 53.44 \\ \hline
\textbf{Eco-MobiLLM} & \textbf{0.10} & 2.86 & \textbf{0.20} & 10.34 & 0.01 & \textbf{2.86} & \textbf{10.34} \\
\hline
\end{tabular}
\end{table*}

\subsubsection{Impact of Side-Network Capacity on Fine-Tuning Performance}
\label{app:capacity_ablation}

To investigate whether the comparable fine-tuning performance of different ablation configurations is attributable to side-network capacity saturation, we further evaluate Eco-MobiLLM and its ablation variants with different internal low-rank hidden sizes of the trainable side-network modules. Specifically, we vary the hidden size $r$ among 16, 64, 128, and 256, with $r=64$ being the default setting in the main experiments. We conduct these experiments using OPT-1.3B on DialogSum and report ROUGE-1, ROUGE-2, and ROUGE-L scores.

As shown in Table~\ref{tab:capacity_ablation}, increasing the hidden size from 16 to 128 generally improves fine-tuning performance, while the results at hidden sizes 128 and 256 remain relatively stable. Across all four hidden sizes, Eco-MobiLLM and its ablation variants achieve comparable performance, with no noticeable performance degradation after introducing TS and TSTP. This consistency is observed even at smaller hidden sizes, where increasing the side-network capacity still leads to clear performance improvements. These results indicate that the comparable performance is not due to side-network capacity saturation, and that the proposed TS and TSTP can preserve fine-tuning performance across different hidden sizes.



\subsection{Breakdown of Per-Batch Execution Time}
\label{app:time_breakdown}

Table~\ref{table:execution_time} compares the time costs, including communication time \( t_1 \), on-device computation time \( t_2 \), and server-side processing time \( t_3 \), for forward and backward passes across different approaches. Split learning based methods execute all the stages sequentially and result in cumulative latency of the three parts.  While MobiLLM supports device-server parallel computation, its heavy activation transmission burden dominates its total LLM FT latency, especially when processing long sequences or large batches. Unlike those designs, Eco-MobiLLM alleviates this bottleneck through token selection, reduces per-batch data transmission and makes transmission latency negligible. For the on-device computing time, both MobiLLM and Eco-MobiLLM remove backward propagation workloads from the mobile devices, which only need to conduct forward inference computations. That helps to simplify on-device operations, making MobiLLM and Eco-MobiLLM naturally compatible with emerging inference accelerators tailored for mobile devices (e.g., mobile TPUs). As shown in Table~\ref{table:execution_time}, Eco-MobiLLM consistently achieves the shortest per-batch execution time on both evaluated devices, demonstrating superior training efficiency.

\subsection{Privacy Discussion on Activation Transmission}\label{app:Privacy}

Eco-MobiLLM focuses on improving the efficiency and deployability of mobile LLM fine-tuning. Eco-MobiLLM involves the transmission of intermediate activations during training. This is aligned with the privacy assumptions commonly adopted in split learning (SL) related architectures~\citep{gupta2018distributed} and federated learning (FL)~\citep{mcmahan2017communication} systems, where devices transmit gradients/activations instead of raw user text data.

However, when data security is strictly considered, prior studies~\citep{zhu2019deep,geiping2020inverting,vepakomma2020nopeek} have shown that gradients or activations, as representations derived from local data, may still be used to infer original input information. To mitigate such risks, existing works~\citep{kairouz2021distributed,stevens2022efficient} have proposed additional protection mechanisms, such as differential privacy (DP) mechanisms~\citep{dwork2006differential} and secure multiparty computation (SMC)~\citep{du2001secure}. DP mechanisms can be applied by adding calibrated Gaussian or Laplace noise to uploaded gradients, activations, outputs, or objective functions, thereby reducing the risk of privacy leakage from transmitted information. SMC can be used as a complementary cryptographic mechanism to further protect the communication and computation process. These protection mechanisms can also be applied to Eco-MobiLLM’s activation transmission.

Compared with conventional SL/FL-style methods, Eco-MobiLLM has a favorable property from the perspective of privacy-loss composition. Conventional SL/FL-style methods usually require multiple communication rounds. If noisy gradients, activations, or model updates are released in each round, the total privacy loss accumulates with the number of releases. To satisfy the same overall privacy budget, the privacy cost of each release must be more tightly controlled, which typically requires stronger noise and may affect model performance. In contrast, Eco-MobiLLM uploads the selected activations for each local sample only once. The server then repeatedly trains the side-network using the cached activations, without requiring devices to upload new activations or gradients in every round. Therefore, Eco-MobiLLM reduces repeated data-dependent release events and can mitigate privacy-loss accumulation. Under the same overall privacy budget, it has the potential to use less aggressive noise calibration and reduce the negative impact of privacy-preserving noise on training performance.

\subsection{Discussion on Token Selection and Activation Caching}
\label{app:discuss ts}

Token selection and data caching are widely used as general concepts. However, our contribution does not lie in these concepts themselves, nor does Eco-MobiLLM simply apply existing techniques. Instead, our contribution is to exploit the architectural properties of server-assisted side-tuning and develop tailored token-selection and activation-caching mechanisms to improve communication and computing efficiency.

First, token selection methods used in areas such as context compression~\cite{li2025prompt} usually rely on a token-importance criterion to identify and retain important tokens, mainly to reduce inference or context-processing costs. Eco-MobiLLM follows a different design. As explained in Sec.~\ref{sec: 4.3}, the side-network adapters process tokens independently. This allows us to move the token selection step before activation transmission and send only the activations at positions used for supervised loss calculation. No semantic token-importance estimation is required. Therefore, our token selector is specifically designed for the side-tuning architecture and the device-server communication bottleneck, rather than being a direct application of existing token pruning or compression methods.

Second, caching is widely used in settings such as KV caching~\cite{shi2024keep}. For example, KV caching reuses the unchanged key/value representations of previous tokens during autoregressive generation to avoid repeated inference computation. In Eco-MobiLLM, the two-stage training paradigm uses a different property: the device-side backbone remains frozen during fine-tuning. As a result, the intermediate activations of the same sample remain unchanged across epochs. We therefore cache these activations on the server to avoid repeated device-side computation and communication.

\end{document}